\documentclass[10pt]{article} 

\usepackage[preprint]{rlj} 

\usepackage{amssymb}            
\usepackage{mathtools}          
\usepackage{mathrsfs}           
\usepackage{graphicx}           
\usepackage{subcaption}         
\usepackage[space]{grffile}     
\usepackage{url}                
\usepackage{lipsum}             
\usepackage[nohypertypes={acronym,main}]{glossaries}
\usepackage{amsthm}
\usepackage{wrapfig}
\makeglossaries

\title{Post-Hoc Robustness for Model-Based Reinforcement Learning}

\setrunningtitle{Post-Hoc Robustness For MBRL}

\author{Siemen Herremans\textsuperscript{1}, Ali Anwar\textsuperscript{1, $\dagger$}, Siegfried Mercelis\textsuperscript{1,$\dagger$}}   

\emails{\{firstname.lastname\}@uantwerpen.be}

\affiliations{
$^{1}$\textbf{IDLab - University of Antwerp, imec}

\par 
$^\dagger$ Equal supervision.
}

\contribution{
    We introduce a methodology to improve the robustness of model-based RL algorithms at inference time (post-hoc).
    }
    {
    Many works focus on modifying the training process to make algorithms more robust, often by performing adversarial training in robust Markov decision processes \citep{wiesemann2013robust}. We recognize that having access to a learned model (as is the case in model-based RL) opens the door to improving the robustness post-hoc, without any training modifications.
    }
\contribution{
    We recognize that the methodology can push a learned model out of distribution and show that a methodology from the offline-RL domain can tackle this.
    }
    {
    As our method modifies predicted transitions, it risks exploiting the model in out-of-distribution areas \citep{janner2019trust}. Therefore, we employ a method from offline RL \citep{morel} and demonstrate the benefit. 
    }
\contribution{
    We experimentally demonstrate the potential of the method by evaluating the robustness on \gls{mujoco} environments and performing ablations.
    }
    {
    Following other works \citep{zhou2024natural,rajeswaran2017epopt,pinto2017robust}, we evaluate the robustness under disturbed environnment parameters in Gymnasium \gls{mujoco}.  
    }


\summary{To improve the real-world applicability of reinforcement learning (RL), the field of adversarially robust RL studies how to train agents under adversarial environment perturbations. In this setting, a protagonist agent optimizes a policy under environmental perturbations from an adversary, resulting in a zero-sum Markov game. When adversarially robust RL is combined with model-based RL, the adversary can target a learned transition model instead of the training environment. Extending this idea, this work introduces post-hoc robustification of deep RL agents at inference time. By using the learned model in combination with a trained nominal policy, our approach performs a robust policy improvement step. The goal is to improve robustness without any additional training of neural networks. Specifically, we utilize model-predictive control under adversarial rollouts, which are approximated via projected gradient descent within a bounded uncertainty set. Furthermore, these offline rollouts are performed while considering and mitigating out-of-distribution issues. The proposed methodology is validated by demonstrating significant improvements in robustness when the algorithm is evaluated in perturbed Gymnasium MuJoCo environments, while considering the computational limitations of the post-hoc inference setting.
}

\begin{document}

\newacronym{rl}{RL}{reinforcement learning}
\newacronym{mbrl}{MBRL}{model-based reinforcement learning}
\newacronym[plural=MDPs, longplural={Markov decision processes}]{mdp}{MDP}{robust Markov decision process}
\newacronym{s2r}{sim2real}{simulation-to-reality}
\newacronym{mbpo}{MBPO}{model-based policy optimization}
\newacronym{kl}{KL}{Kullback-Leibler}
\newacronym{rmbpo}{RMBPO}{robust model-based policy optimization}
\newacronym{sac}{SAC}{Soft Actor-Critic}
\newacronym{iqm}{IQM}{interquartile mean}
\newacronym{rarl}{RARL}{robust adversarial reinforcement learning}
\newacronym{rnac}{RNAC}{robust natural actor-critic}
\newacronym{ewok}{EWoK}{Estimate the Worst Kernel}
\newacronym{qarl}{QARL}{Quantal Adversarial RL}
\newacronym{neld}{MixedNE-LD}{mixed Nash Equilibrium via Langevin dynamics}
\newacronym{mujoco}{MuJoCo}{Multiple Joint Control}
\newacronym{ds}{DS}{double sampling}
\newacronym{ipm}{IPM}{integral probability metric}
\newacronym{tv}{TV}{total variation}
\newacronym{rmdp}{RMDP}{robust Markov decision process}
\newacronym{mle}{MLE}{maximum likelihood estimation}
\newacronym{mdmm}{MDMM}{modified method of differential multipliers}
\newacronym{dmc}{DMC}{Deepmind Control Suite}
\newacronym{utd}{UTD}{update-to-data}

\newacronym{mpc}{MPC}{model-predictive control}
\newacronym{pgd}{PGD}{projected gradient descent}
\newacronym{ood}{OOD}{out-of-distribution}
\newacronym{s2s}{sim2sim}{simulation-to-simulation}
\newacronym{otc}{OTC}{optimal transport cost}
\newacronym{mppi}{MPPI}{model-predictive path integral}

\maketitle  

\begin{abstract}
To improve the real-world applicability of reinforcement learning (RL), the field of adversarially robust RL studies how to train agents under adversarial environment perturbations. In this setting, a protagonist agent optimizes a policy under environmental perturbations from an adversary, resulting in a zero-sum Markov game. When adversarially robust RL is combined with model-based RL, the adversary can target a learned transition model instead of the training environment. Extending this idea, this work introduces post-hoc robustification of deep RL agents at inference time. By using the learned model in combination with a trained nominal policy, our approach performs a robust policy improvement step. The goal is to improve robustness without any additional training of neural networks. Specifically, we utilize model-predictive control under adversarial rollouts, which are approximated via projected gradient descent within a bounded uncertainty set. Furthermore, these offline rollouts are performed while considering and mitigating out-of-distribution issues. The proposed methodology is validated by demonstrating significant improvements in robustness when the algorithm is evaluated in perturbed Gymnasium MuJoCo environments, while considering the computational limitations of the post-hoc inference setting.
\end{abstract}

\section{Introduction}
\begin{figure}
    \centering
    \includegraphics[width=0.99\linewidth]{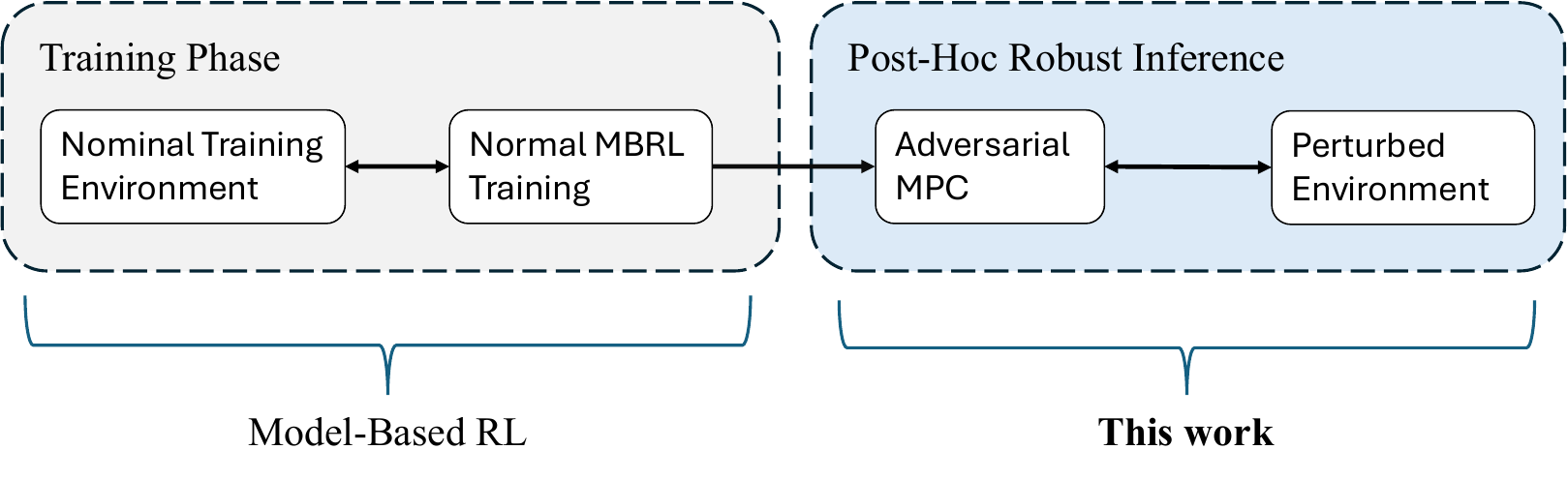}
    \caption{This work considers post-hoc robustification, located at inference time, without modifiying the training process of the agent.}
    \label{fig:overview}
\end{figure}

Improving the robustness of \gls{rl} is a challenge that is tackled from multiple perspectives. One important reason to improve the robustness of \gls{rl} is to handle \gls{s2r} or \gls{s2s} settings, where an \gls{rl} agent must be robust enough to still perform well in deployment environments that are slightly different from the training environment. A possible approach to tackle this problem is domain randomization, which tries to make the agent generalize by adding noise to its training environment. While useful in many scenarios, this method has the consideration that it still optimizes for the average case, and not the worst case. A common framework to optimize for worst-case scenarios is an \gls{rmdp}. This formulation considers that the agent is operating in an uncertainty set of possible environments, where it is optimized to maximize its performance in the adversarially selected (worst-case) environment of that set. In a deep \gls{rl} setting, the \gls{rmdp} formalism is often employed by adversarial algorithms, where the adversary can apply perturbations to the training process of the agent, such as to the environment \citep{janner2019trust, qarl, wang2024bring}, the action space \citep{tessler2019action} or to a learned model \citep{queeney2024optimal,herremans2025robust}. This paper, however, aims to investigate the possibility of improving the robustness of a pre-trained \gls{mbrl} agent at inference time, hence entirely omitting any modifications to the agent's training process. We denote this setting as post-hoc robustness, as shown in Fig. \ref{fig:overview}. This, in principle, allows one to define the uncertainty set and the robustness/optimality trade-off in a later phase when deemed necessary. In addition, by leaving the training process unaltered, there is no performance decay or instability introduced, typically associated with adversarial training in an \gls{rmdp}.

Our work is inspired by the \gls{rmdp} setting and proposes an algorithm that improves the robustness of a previously-trained \gls{mbrl} agent at inference time without any re-training. Our approach performs \gls{mpc} on the learned transition model, while introducing two crucial modifications. Firstly, the model-predictive rollouts are performed in an adversarial manner. More precisely, every predicted state is optimized via \gls{pgd} to minimize its predicted value, while remaining within a radius of the original prediction. In contrast to a true \gls{rmdp} setting, we do not have access to the robust value function and are therefore pessimistic with regards to the nominal value function. Secondly, as we are locally updating the policy without environment interaction, care must be taken to avoid \gls{ood} usage of the learned transition model. We tackle this issue by truncating the rollout depth based on epistemic uncertainty estimates, as introduced by \cite{morel}. 

Our \textbf{main contributions} are firstly \textbf{(1)}, proposing a practical algorithm to improve robustness in a post-hoc manner at inference time. Secondly \textbf{(2)}, we evaluate the empirical performance of our methodology on limited-complexity \gls{mujoco} benchmarks under single and simultaneous parameter perturbations. In addition, ablations are performed to isolate where the improvement of our method lies. Finally \textbf{(3)} we also make sure to evaluate the computational cost/duration on different hardware, to align with the inference-time setting.
The remainder of this work will first highlight relevant background to our approach. Then, our methodology is described. Subsequently, the results evaluate the improvement in robustness of our method to the unmodified (baseline) \gls{mbpo} \citep{janner2019trust} algorithm in \gls{mujoco} \citep{todorov2012mujoco} control environments. Finally, we discuss current limitations of the method and outline future research directions.

\section{Background}
This section briefly introduces the notational conventions that we use throughout this work to describe \gls{mbrl} and a \gls{mdp}. Subsequently, we provide a background on \gls{rmdp}s and the \gls{otc} uncertainty set. 

\subsection{Model-Based Reinforcement Learning in a Markov Desision Process}
\label{sec:mbrl}

\Gls{mdp}s are defined by a tuple $(\mathcal{S}, \mathcal{A}, P, R, \gamma, \rho_0)$. In this definition, $\mathcal{S}$ denotes the state space and $\mathcal{A}$ denotes the action space. Transition probabilities are denoted by $P(s'|s,a)$ when in state $s$ and performing action $a$. Furthermore, a reward function $R(s,a)$ defines the expected reward of these transitions. To balance between short-term rewards and long-term returns, a discount factor $\gamma$ is used. Finally, an initial state distribution $\rho_0(s)$ defines the likelihood that the agent will start in a given state $s$. \Gls{rl} aims to converge to the most optimal policy $\pi^*(a|s)$ from the set of available policies $\Pi$. This policy maximizes the expected sum of discounted rewards:

\begin{equation}
    \pi^* = \mathop{\arg\max}_{\pi\in\Pi} \mathbb{E}_{\pi,P} \left[ \sum_{t=0}^\infty \gamma^t r_t \right].
\end{equation}

We will also be making use of the value function $V^\pi$, defined as:
\begin{equation}
    V^{\pi}(s) = \mathbb{E}_{\pi, P} \left[ \sum_{t=0}^{\infty} \gamma^t r_t \mid s_0 = s \right].
    \label{eq:value_function}
\end{equation}

In contrast with model-free approaches, \gls{mbrl} employs a learned transition model, which we denote by $p_\theta(s',r|s,a)$, that approximates the true transition and reward functions from collected data. This model is then used to simulate future states and rewards, reducing the number of required interactions with the real \gls{mdp}. This learned transition model can then be used to generate approximate environment transitions. It is possible to perform rollouts on this model, which recursively use this learned transition model to generate a sequence of transitions. These rollouts are useful for training a policy, but in this work, we will additionally use $p_\theta$ to optimize a local policy distribution $\pi_{s_t}$ in a given state $s_t$ at inference time.

\subsection{Robust Markov Decision Processes}
\label{sec:rmdp}

When training an agent that is expected to be acting under perturbations at inference time, it is useful to consider an uncertainty set of \gls{mdp}s, where the transition distribution lies within that set, $P(s'| s,a) \in \mathcal{P}^{s,a}$. When the agent acts under transition distributions that are adversarially selected from that set, the optimization setting can be denoted as an \gls{rmdp} \citep{wiesemann2013robust}. We follow a common assumption in literature where we only consider $sa$-rectangular uncertainty sets, to keep the problem setting tractable \citep{wiesemann2013robust,wang2024bring,zhou2024natural}. This assumption ensures that an uncertainty set in a $(s,a)$-pair is always independent from the uncertainty set in other $(s,a)$-pairs. Consequently this allows the global uncertainty set $\mathcal{P}$ to be constructed as the Cartesian product of independent marginal uncertainty sets $\mathcal{P}^{s,a}$. In an \gls{rmdp}, the robust objective $J_{robust}(\pi)$ is the objective function in the worst-case \gls{mdp} of a predefined uncertainty set, formalized in Eq.~\ref{eq:objective}.

\begin{equation}
J_{robust}(\pi) = \min_{P \in \mathcal{P}} \mathbb{E}_{\pi,P} \left[ \sum_{t=0}^\infty \gamma^t r_t \right].
\label{eq:objective}
\end{equation}

Similar to a normal \gls{mdp}, the robust value function $V^\pi_{robust}$, can be defined as:
\begin{equation}
    V^{\pi}_{robust}(s) = \min_{P \in \mathcal{P}} \mathbb{E}_{\pi, P} \left[ \sum_{t=0}^{\infty} \gamma^t r_t\mid s_0 = s \right].
    \label{eq:robust_value_function}
\end{equation}

We refer to the problem of finding the $P \in \mathcal{P}$ as the \emph{inner problem}. Considering this objective, robust \gls{rl} aims to approximate the optimal robust policy $\pi^*_\mathcal{P}$, that maximizes the worst-case performance within the uncertainty set. We refer to this problem as the \emph{outer problem}, described in Eq. \ref{eq:robust_policy}.

\begin{equation}
\pi^{*}_{\mathcal{P}} = \mathop{\arg\max}_{\pi \in \Pi} J_{robust}(\pi)
\label{eq:robust_policy}
\end{equation}

The combination of the inner and outer problem can be viewed as a zero-sum Markov game \citep{rigter2022rambo, pinto2017robust}, where one player optimizes the policy, to maximize the expected return, whilst the other player tries to find $P^* \in \mathcal{P}$, which minimizes the expected return.

\subsection{Optimal Transport Cost Uncertainty set}
\label{sec:otc_uncertainty}
To define an \gls{rmdp}, the uncertainty sets $\mathcal{P}^{s,a}$ must be constructed. This is commonly done by allowing all transition distributions $P(\cdot|s,a)$ that are within a certain distance $\epsilon_{s,a}$ of a nominal distribution $\bar{P}(\cdot|s,a)$. A commonly used example of such an uncertainty set uses the \gls{kl} divergence \citep{hu2013kullback, wang2024bring, shi2024distributionally}. Recently, \cite{queeney2024optimal} have demonstrated that it is possible to construct an uncertainty set, based on the \gls{otc} which allows to define the size of the uncertainty set directly in terms of state perturbations instead of via measures on the difference in transition probability vectors. For any continuous function $d_{s,a}: \mathcal{S} \times \mathcal{S} \rightarrow \mathbb{R^{+}}$, where $d_{s,a}(s',s')$ is always zero for any $s'$, the \gls{otc} is defined as follows:

\begin{equation}
\textnormal{OTC}_{d_{s,a}} ( \bar{P}(\cdot|s,a), P(\cdot|s,a) ) = \inf_{\nu \in \Gamma \left( \bar{P}(\cdot|s,a), P(\cdot|s,a) \right)} \int_{\mathcal{S} \times \mathcal{S}} d_{s,a} ( \bar{s}', s' ) \textnormal{d}\nu ( \bar{s}', s' ),
\end{equation}

where $\Gamma ( \bar{P}(\cdot|s,a), P(\cdot|s,a) )$ is the set of all possible ways to map the probability mass between $\bar{P}(\cdot|s,a)$ and $P(\cdot|s,a)$. Using this metric, \cite{queeney2024optimal} defines the \gls{otc} uncertainty set as follows:

\begin{equation}
    \mathcal{P}^{s,a}_\textnormal{OTC} = \left\lbrace P(\cdot|s,a) \in \Delta_\mathcal{S} \mid \textnormal{OTC}_{d_{s,a}}(\bar{P}(\cdot|s,a), P(\cdot|s,a)) \leq \epsilon_{s,a} \right\rbrace,
\label{eqn:otc_set}
\end{equation}
where $\epsilon_{s,a}$ is a constant that can be chosen to define the size of the uncertainty set, and $\Delta_\mathcal{S}$ denotes the probability simplex over $\mathcal{S}$. This formulation is very broad as it allows us to pick from a broad range of functions $d_{s,a}(\bar{s}',s')$ that compare the next states. Also, the \gls{otc} uncertainty set is not necessarily restricted to transition distributions that must have overlapping support, such as when using the \gls{kl} divergence. In general, this allows the uncertainty set to be defined in expected perturbations on the states themselves, without needing the probability vectors. If the transitions are deterministic, a single measure $d_{s,a} ( \bar{s}', s' )$ can fully define $\mathcal{P}^{s,a}_{\textnormal{OTC}}$. As we are assuming (s,a)-rectangularity, we can denote the global \gls{otc} uncertainty set as $\mathcal{P_{\textnormal{OTC}}}$.

\section{Methodology}


Our method proposes a post-hoc robustification method on a \gls{mbrl} agent that was trained in a nominal environment $\bar{P}$. We propose to use \gls{mpc} over model-based pessimistic rollouts at each step in a (perturbed) evaluation environment. Crucially, given the post-hoc setting where further training can not be performed, we do not have access to the robust value function $V^{\pi}_{robust}(s)$, but only to the nominal value function $V^{\pi}(s)$ that was learned during the training phase in the nominal environment. This means that we can not truly optimize the robust objective from Eq. \ref{eq:objective}. However, we propose to perform value estimates using rollouts as follows (described in a recursive manner):
\begin{equation}
    V_k^{rollout}(s_k) = r(s_k, \pi(s_k)) + \gamma \, V_{k+1}^{rollout}(s_{k+1}), \ \text{where}\ s_{k+1} \sim \underset{P \in \mathcal{P}}{\arg\min} \, \mathbb{E}_{s' \sim P(\cdot|s,a)}
    \left[ V^\pi(s') \right],
    \label{eq:adversarial_rollout}
\end{equation}
where $k = 0, \ldots, N-1$ and $N$ is the rollout horizon. At the final timestep N, we use $V^\pi(s_N)$. Note that increasing the rollout horizon $N$ yields increasingly pessimistic estimates,
but this does not guarantee that the true robust value is being approached. In the case where $N=1$, the rollouts become a single iteration of the robust Bellman operator \citep{iyengar2005robust}. However, this strict relation to the Robust bellman operator disappears for other values of $N$. Despite this drift, our experiments show empirically that a larger $N$ can significantly improve robustness. To tackle the outer problem, by optimizing a local policy distribution in a given state, we define the MPC policy as:
\begin{equation}
    \pi_{s}^{MPC} = \underset{\pi_s}{\arg\max} \, \mathbb{E}_{\pi_s} \left[ V^{rollout}(s) \right].
    \label{eq:mpc_policy}
\end{equation}

This policy is approximated by a version of \gls{mppi} \citep{williams2015model}, which was adapted by \cite{hansen2024tdmpc} to work in the context of \gls{mbrl} agents. As we are in a deep learning setting, the learned value function $V^\pi_\psi \approx V^\pi$ is used and the \gls{mppi} algorithm uses the learned policy $\pi_\psi$ as its initialization. This leaves the question of how to compute $\underset{P \in \mathcal{P}}{\arg\min} \, \mathbb{E}_{s' \sim P(\cdot|s,a)} \left[ V^\pi(s') \right]$, which will be described in Sec. \ref{subsec:value_minimization}. Furthermore, Sec. \ref{subsec:uncertainty reduction} describes how to deal with the increase in epistemic uncertainty that can arise from performing long rollouts on a learned transition model.


\subsection{Model-Based pessimistic transitions}
\label{subsec:value_minimization}

We choose $\mathcal{P}$ to be $\mathcal{P}_{otc}$ as this allows to define the uncertainty on states instead of transition probability vectors. More specifically, we choose $d_{s,a}(\bar{s}',s')=\left | \left | \bar{s}' - s'\right | \right |_2$, where $\bar{s}'$ is the next state that is predicted by the learned nominal transition model $p_\theta$, and $s'=\bar{s}'+\delta$ is the perturbed next state. During model-based unrolls, we perturb the predicted next state $s'$ by applying normalized \gls{pgd} iterations with step size $\alpha$ on the state perturbation. The projection step ensures that the state disturbance will remain within an L2 ball with a limited radius, and therefore that we remain within an \gls{otc} uncertainty set $\mathcal{P}_{otc}$, defined around the approximate model $p_\theta$. Formally, the disturbance is updated in steps to minimize the learned value function $V^\pi_\psi(s')$ as follows:

\begin{equation}
    \delta \leftarrow \mathrm{Proj}_{\|\delta\|_2 \le r^2}\left(\delta-\alpha \,\frac{\nabla_{(s' + \delta)}{V_\psi(s' + \delta)}} {\|\nabla_{(s' + \delta)} V_\psi(s' + \delta)\|_2}\right),
\end{equation}

where $\mathrm{Proj}_{\|\delta\|_2 \le r^2}$ denotes the function that projects the updated perturbation back into the uncertainty set, if needed. This projection is described by: 
\begin{equation}
    \mathrm{Proj}_{\|\delta\|_2 \le r^2}(\delta)=\delta \cdot \min\!\left(1,\; \frac{r^2}{\|\delta\|_2}\right),
\end{equation}

where $r^2$ is the radius of the uncertainty set. In our experiments, one to three \gls{pgd} iterations were always enough to converge to a (local) optimum.

\subsection{Epistemic uncertainty reduction}
\label{subsec:uncertainty reduction}

\begin{wrapfigure}{r}{0.45\textwidth} 
    \centering
    \vspace{-20pt} 
    \includegraphics[width=\linewidth]{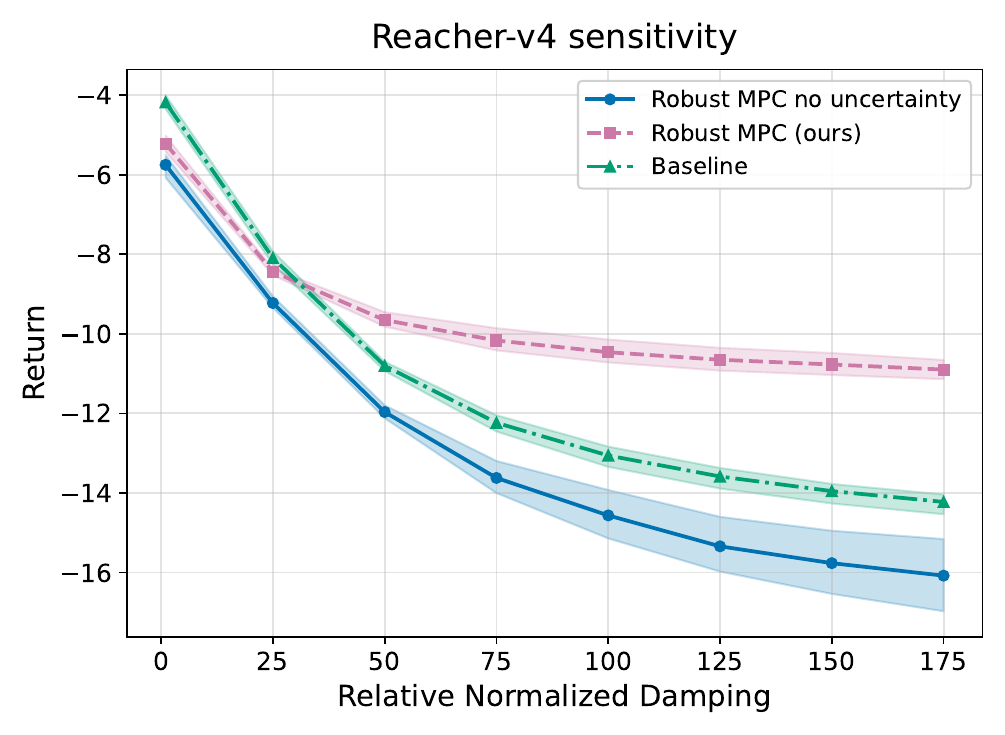}
    \caption{Robust MPC ablation on Reacher-v4.}
    \label{fig:varpen_ablation}
    \vspace{-15pt} 
\end{wrapfigure}
In the proposed method, model-based rollouts are performed on a learned model $p_\theta$. Although we are not in a full offline-RL setting (as we start from on-policy rollouts), there are two aspects that cause \gls{ood} problems during model unrolls. Firstly, because we add perturbations to the predicted transitions to compute $\underset{P \in \mathcal{P}}{\arg\min} \, \mathbb{E}_{s' \sim P(\cdot|s,a)} \left[ V^\pi(s') \right]$. Secondly, during \gls{mpc} iterations, the updated policy $\pi^{MPC}_{s}$ can deviate from the learned policy $\pi_\psi(s)$. This is of course the goal, as we want to robustify the policy, but it also leads to \gls{ood} usage of the learned transition model. \cite{janner2019trust} have shown that both of these aspects increase the upper bound on the model error. Therefore, we follow we follow \cite{morel} in using the ensemble variance to truncate model rollouts, starting from \gls{mbpo} \citep{janner2019trust} as a base. Concretely, using a bootstrapped ensemble of learned transition models $\{ p_{\theta_1}, p_{\theta_2}, \ldots \}$, we compute the ensemble discrepancy as $epistemic(s,a)=\max_{i, j} \ \left\| p_{\theta_i}(s,a)-p_{\theta_j}(s,a) \right\|_2$. We define a threshold $h$ which is used to truncate the rollouts from Eq. \ref{eq:adversarial_rollout} when $epistemic(s,a) > h$. This method ensures that the learned transition model will not be exploited in areas where the ensembles do not agree on the transition. To verify the necessity and the effectiveness of this addition, we evaluate it in Fig. \ref{fig:varpen_ablation}, where the baseline \gls{mbpo} policy $\pi_\psi$ is compared with our method $\pi^{MPC}_s$, with and without the epistemic uncertainty reduction. It can be seen that our method successfully improves the robustness of the policy in that experiment, but without the epistemic uncertainty reduction, the performance does not even reach baseline level. This confirms the danger of model exploitation and the necessity of using the method from \cite{morel} in this setting.
\label{sec:methodology}

\section{Experiments}

To evaluate our methodology, we pre-train \gls{mbpo} on the Gymnasium \gls{mujoco} Reacher-v4 and Hopper-v4 environments, after which we evaluate the agent under environment perturbations. We use unmodified \gls{mbpo} as a baseline, where $a_t \sim \pi_\psi(s_t)$, and compare it with our post-hoc robustness method, acting with $a_t \sim \pi_{s_t}^{MPC}$. Additionally, a version that uses \gls{mpc} at inference time, but does not perform the distortions of Sec. \ref{subsec:value_minimization}, is also included as an ablation, discussed in Sec. \ref{subsec:ablations}. In the evaluation environments, we follow \cite{pinto2017robust} in perturbing the mass of the torso of the robot, and the friction coefficient. By following the choices of \cite{pinto2017robust}, we avoid the possibility of cherry-picking perturbation parameters where our method performs best. For Reacher-v4, we perturb the damping coefficient and the gear ratio of the robot arm. 
All algorithms are trained using 5 different initial seeds, the results display 95\% bootstrapped confidence intervals as shaded regions. For each individually seeded policy, its return is computed as the average performance over 20 evaluation episodes with a different random initialization.

\subsection{Main Results}
\label{subsec:main_results}
\begin{figure}[t]
    \centering
    \begin{subfigure}[t]{0.5\linewidth}
        \centering
        \includegraphics[width=\linewidth]{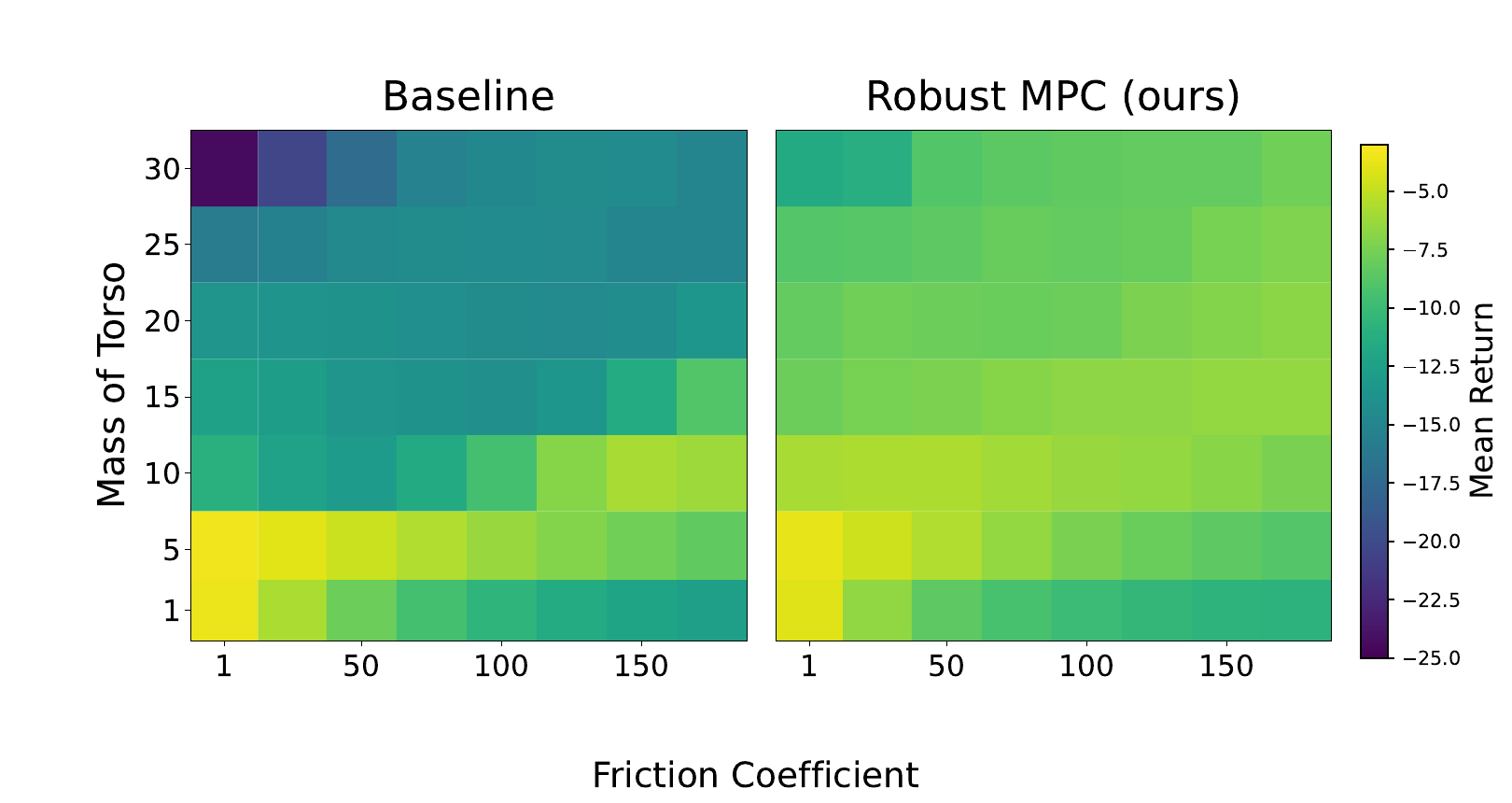}
        \caption{Reacher-v4.}
        \label{subfig:reacher_2d}
    \end{subfigure}\hfill
    \begin{subfigure}[t]{0.5\linewidth}
        \centering
        \includegraphics[width=\linewidth]{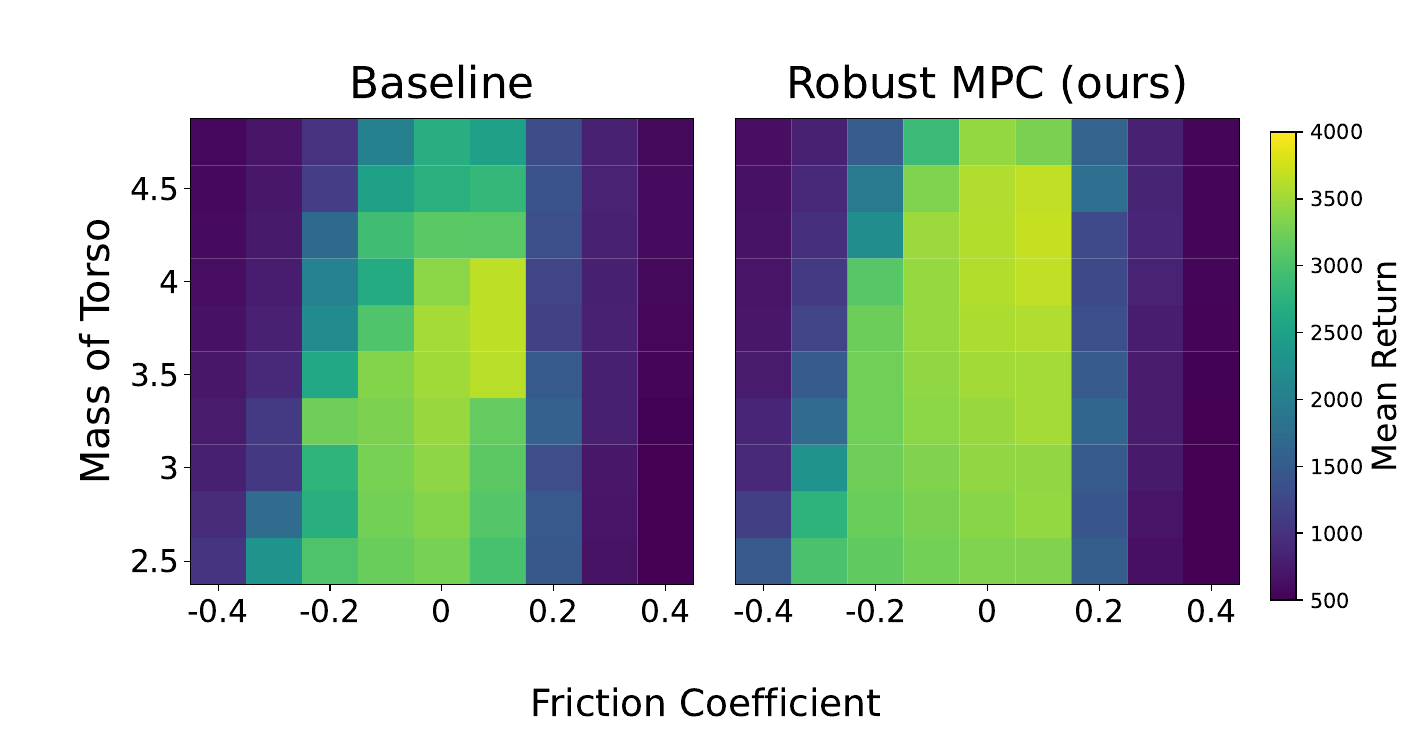}
        \caption{Hopper-v4.}
        \label{subfig:hopper_2d}
    \end{subfigure}
\caption{Evaluation under two simultaneous perturbations.}
\label{fig:2d_results}
\end{figure}

This section investigates the potential of our methodology to improve a pre-trained \gls{mbpo} policy at inference time. Therefore, we compare the performance of the normal \gls{mbpo} policy with a policy that performs robust \gls{mpc} as described in \ref{sec:methodology}. We evaluate the performance both under a single environment parameter distortion and for two simultaneous distortion. Firstly, Fig. \ref{fig:reacher_1d} and Fig. \ref{fig:hopper_1d} demonstrate that our method significantly improves the robustness. In all cases, the robust \gls{mpc} reduces the performance decay under perturbations, confirming the claim that this work can improve robustness. Furthermore, we evaluate the performance under two simultaneous environment perturbations in Fig. \ref{fig:2d_results}, again demonstrating a significant improvement in robustness, compared to the baseline. 
Additionally, as we are operating in an inference-time setting, it is relevant to consider the runtime of the algorithm for each environment step, therefore, we report these timing results in Table \ref{table:compute} for a variety of hardware.

\subsection{Ablations}
\label{subsec:ablations}

\begin{figure}[t]
    \centering
    \begin{minipage}{0.85\linewidth} 
        \centering
        \begin{subfigure}[b]{0.48\linewidth}
            \centering
            \includegraphics[width=\linewidth]{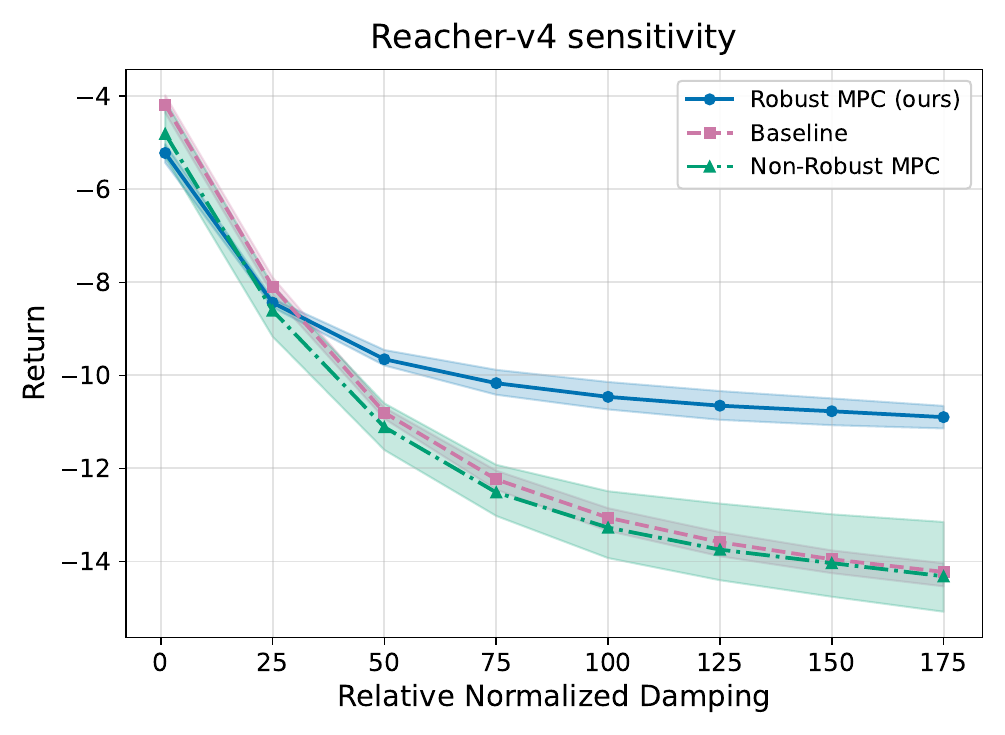}
            \label{subfig:reacher_damping}
        \end{subfigure}\hfill
        \begin{subfigure}[b]{0.48\linewidth}
            \centering
            \includegraphics[width=\linewidth]{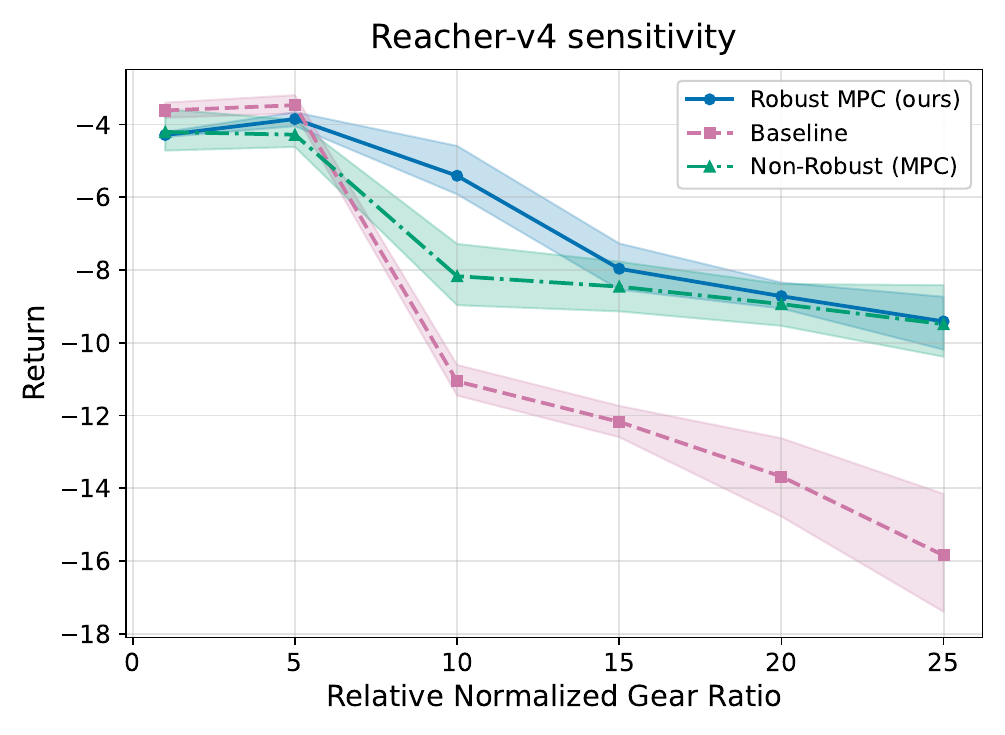}
            \label{subfig:reacher_gear}
        \end{subfigure}
        \caption{Evaluation in Reacher-v4 under a single perturbed environment parameter.}
        \label{fig:reacher_1d}
    \end{minipage}

    \vspace{2em} 

    \begin{minipage}{0.85\linewidth} 
        \centering
        \begin{subfigure}[b]{0.48\linewidth}
            \centering
            \includegraphics[width=\linewidth]{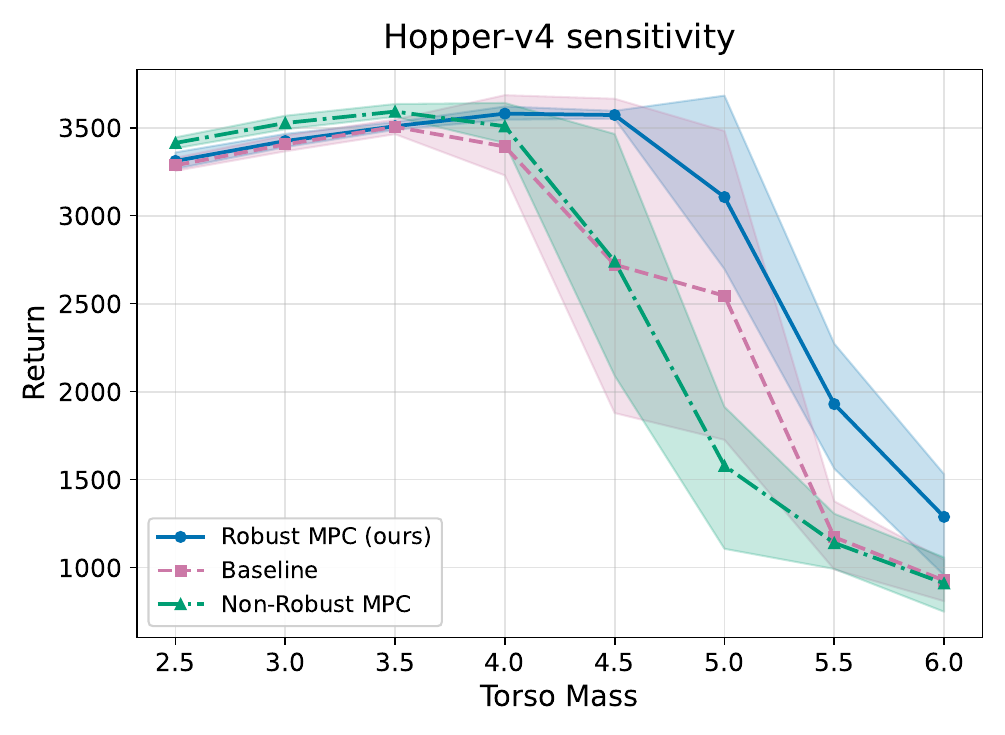}
            \label{subfig:hopper_torso}
        \end{subfigure}\hfill
        \begin{subfigure}[b]{0.48\linewidth}
            \centering
            \includegraphics[width=\linewidth]{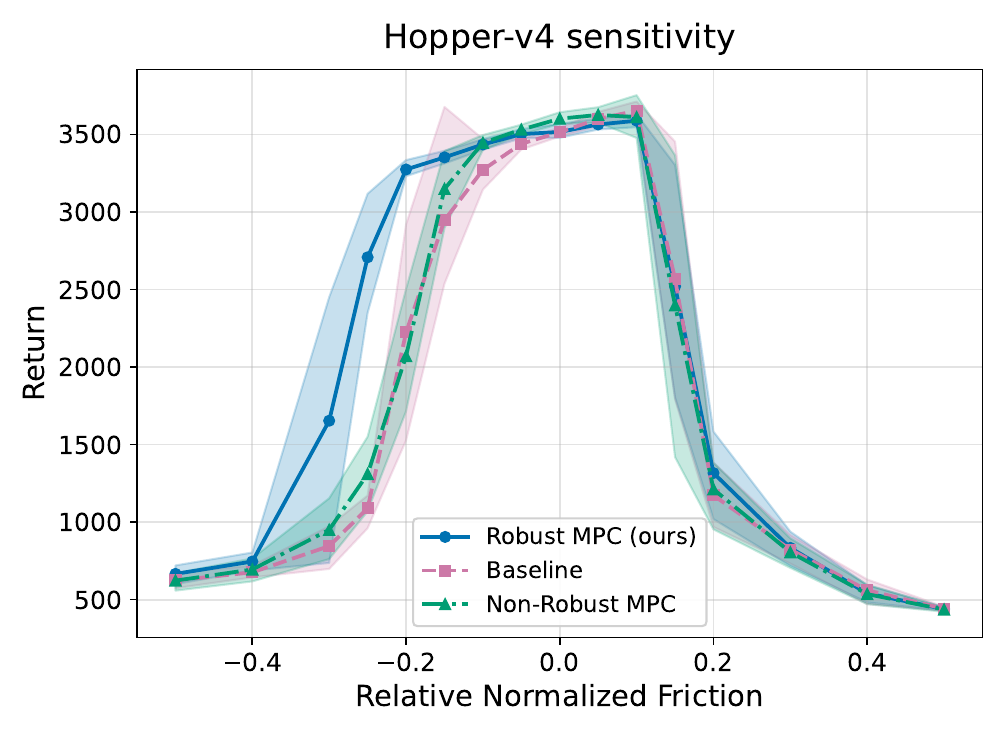}
            \label{subfig:hopper_friction}
        \end{subfigure}
        \caption{Evaluation in Hopper-v4 under a single perturbed environment parameter.}
        \label{fig:hopper_1d}
    \end{minipage}
    
\end{figure}

It is important to verify if the robustness improvement actually comes from the proposed method in Sec. \ref{subsec:value_minimization}. For this, we included an ablation called "Non-Robust MPC" in Fig. \ref{fig:reacher_1d}, and Fig. \ref{fig:hopper_1d}, showing the robustness of performing local \gls{mpc} optimization where the rollouts are performed without the adversarial perturbations of Sec. \ref{subsec:value_minimization}. This non-robust \gls{mpc} approach does still use the \gls{ood} mitigation of \ref{subsec:uncertainty reduction}. These results confirm the need of adversarial perturbations to improve robustness. Often, these perturbations are the only factor contributing to the robustness. However, in Reacher-v4, performing non-robust \gls{mpc} alone already contributes significantly to the robustness. Furthermore, Fig. \ref{fig:ablations} shows the influence of the rollout depth $N$ and the uncertainty set radius $r^2$ for Hopper-v4 under torso mass perturbations. It is shown that longer rollouts can successfully improve the robustness of the method. Also, Fig. \ref{subfig:radius_ablation} demonstrates an improvement in robustness for a larger uncertainty set radius, and that a significant stable region exists ($0.045$, $r=0.06$, $r=0.075$) before the performance decays ($r=0.12$). The same ablations are performed with friction perturbations in Appendix \ref{apx:additional_results}.   

\begin{figure}[htbp]
    \centering
    \begin{subfigure}[h]{0.48\linewidth}
        \centering
        \includegraphics[width=\linewidth]{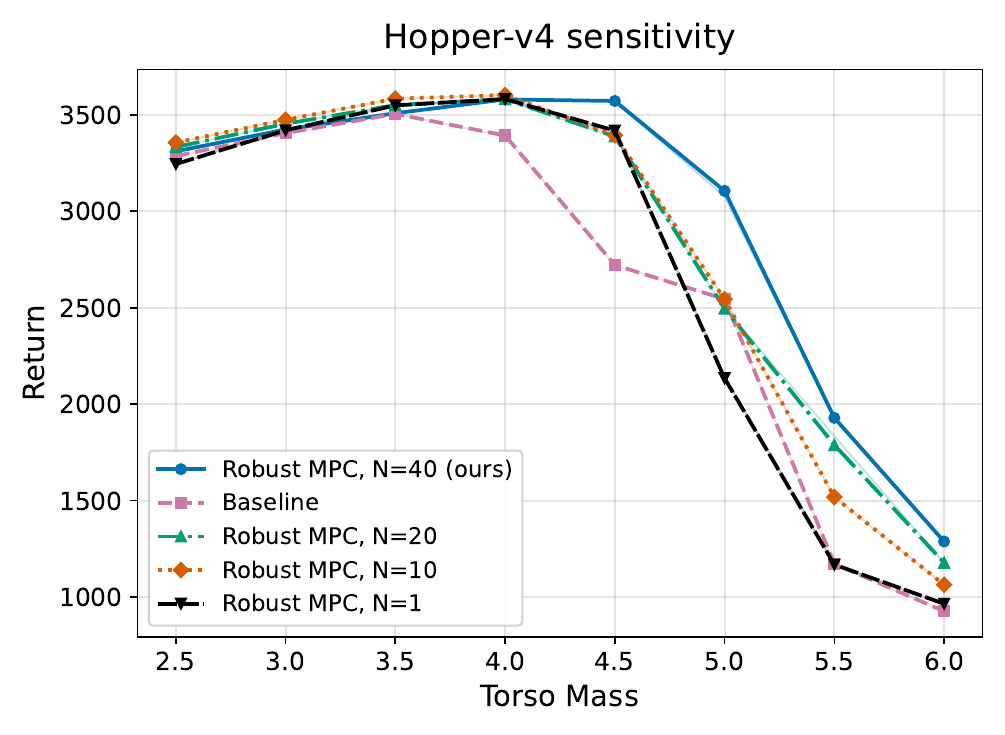}
        \caption{Rollout depth.}
        \label{subfig:rollout_ablation}
    \end{subfigure}\hfill
    \begin{subfigure}[h]{0.48\linewidth}
        \centering
        \includegraphics[width=\linewidth]{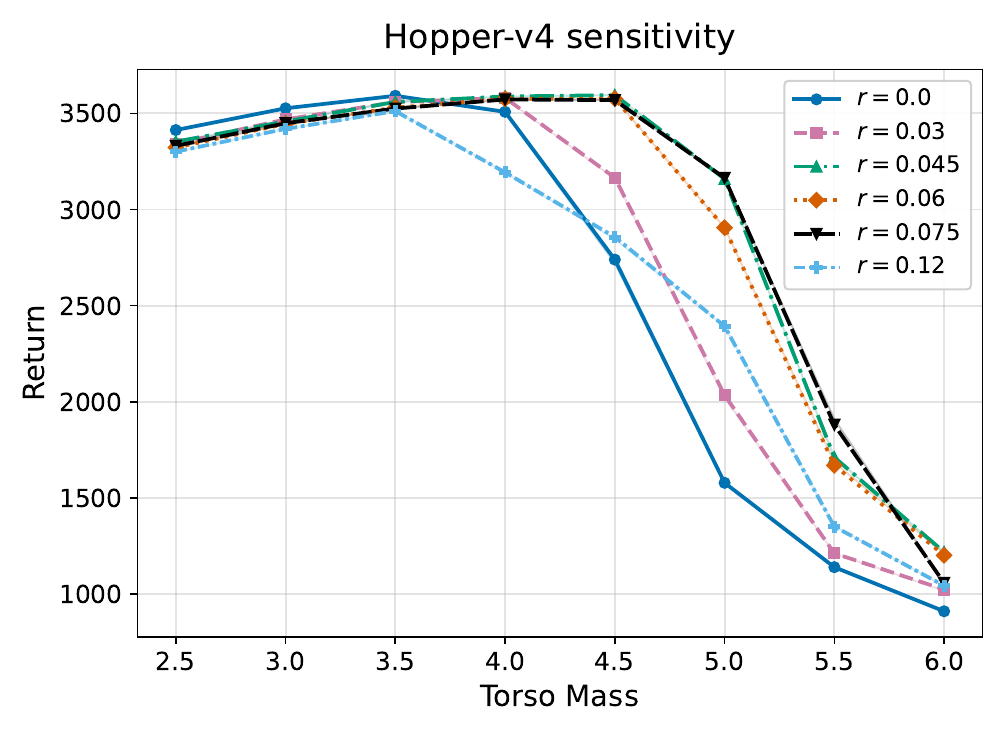}
        \caption{Uncertainty set radius.}
        \label{subfig:radius_ablation}
    \end{subfigure}
    \caption{Influence of the rollout depth and the uncertainty set radius $r^2$. Confidence intervals are omitted for visual clarity.}
\label{fig:ablations}
\end{figure}

\begin{table}[t]
\centering
\begin{tabular}{l c c l}
\hline
\textbf{GPU Type} & \textbf{Time/step Hopper-v4 (N=20)} & \textbf{Time/step Reacher-v4 (N=5)} & \textbf{Data Type} \\
RTX 4090    & 0.09s & 0.015s & bfloat16 \\ 
RTX 2080 Ti & 0.12s & 0.024s & float16  \\  
A6000       & 0.15s & 0.019s & bfloat16 \\    
H200        & 0.09s & 0.014s & bfloat16 \\ \hline
\end{tabular}
\caption{Robust inference duration on different types of GPUs, implemented in Jax \citep{jax}. Including the GPU/CPU memory copies.}
\label{table:compute}
\end{table}

\section{Discussion And Limitations}
In the experiments, our approach substantially increases the robustness against environmental perturbations at inference time. This demonstrates the potential of the method, but also leaves the question of potential issues associated with the adversary minimizing $V^\pi(s')$ instead of $V^\pi_{robust}(s')$. While 1-step rollouts can theoretically be interpreted as a single robust value iteration, this does not hold anymore when performing n-step rollouts. Still, Fig. \ref{fig:ablations} demonstrates that increasing $N$ significantly improves the robustness of our method in Hopper-v4. Despite the lack of formal guarantees, we believe this makes intuitive sense in many control problems, as many transitions that have a high impact under $V^\pi(s')$ will likely also have a high impact under $V^\pi_{robust}(s')$. For example, under Hopper-v4, bringing the robot towards an irrecoverable falling state, will be problematic under both the nominal and the adversarial model. As a clear example, termination causes $V^\pi(s')=0$ and $V^\pi_{robust}(s')=0$, by construction of the environment. However, we do acknowledge that this is a limitation of our method and we believe that understanding this approach more theoretically is an important future work.

\section{Related Works}
A significant body of research exists that investigates adversarial deep \gls{rl} \citep{queeney2024optimal, zhou2024natural, herremans2025robust, qarl, wang2024bring}. These methods focus on adversarial perturbations during training, to reach the Nash equilibrium of an \gls{rmdp}. Some of these works also use a projected gradient to perturb states to minimize the value function \citep{zhang2020robust, NEURIPS2020_f0eb6568}. Furthermore, contextual \gls{rl} can be used to improve the runtime robustness of \gls{rl} by adaptively considering the context at deployment \citep{benjamins2023contextualize, chen2022an, pmlr-v119-lee20g, clavera2018learning, queeney2025gram}. In contrast to our setting, contextual \gls{rl} requires access to the context that the agent is operating in at training time, and needs a variety of contexts to train in. This context can later be used at deploy time to adapt the behavior accordingly. 

\section{Conclusion and Future Work}
This work introduces a methodology to improve the robustness of an \gls{mbrl} algorithm at inference time. This is achieved by performing \gls{mpc} with pessimistic transitions on a pre-trained \gls{mbpo} model, while considering \gls{ood} issues that arise in this setting. Results demonstrate a significant improvement in robustness in limited-complexity \gls{mujoco} environments, without needing to adapt the training process of the agent. Additionally, we show this can be done with a limited amount of compute to per step. Although our method shows potential, future work is needed to investigate how well this method can scale to larger and more complex problems. In addition, it might be interesting future work to investigate the effect of using nominal value minimization instead of robust value minimization in this setting.

\appendix





\subsubsection*{Acknowledgments}
\label{sec:ack}
This work was supported by the Research Foundation Flanders (FWO) under Grant Number 1SHAI24N.


\bibliography{main}
\bibliographystyle{rlj}

\beginSupplementaryMaterials
\section{Additional results}
\label{apx:additional_results}

\begin{figure}[h]
    \centering
    \begin{subfigure}{0.48\linewidth}
        \centering
        \includegraphics[width=\linewidth]{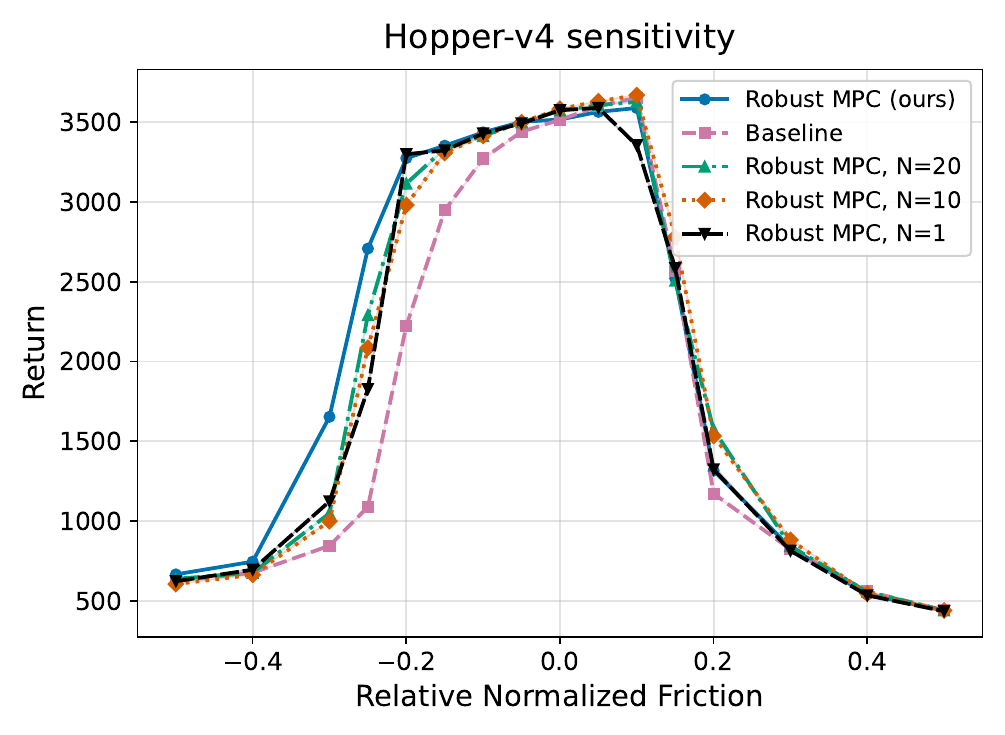}
        \caption{Rollout depth.}
        \label{subfig:rollout_ablation}
    \end{subfigure}\hfill
    \begin{subfigure}{0.48\linewidth}
        \centering
        \includegraphics[width=\linewidth]{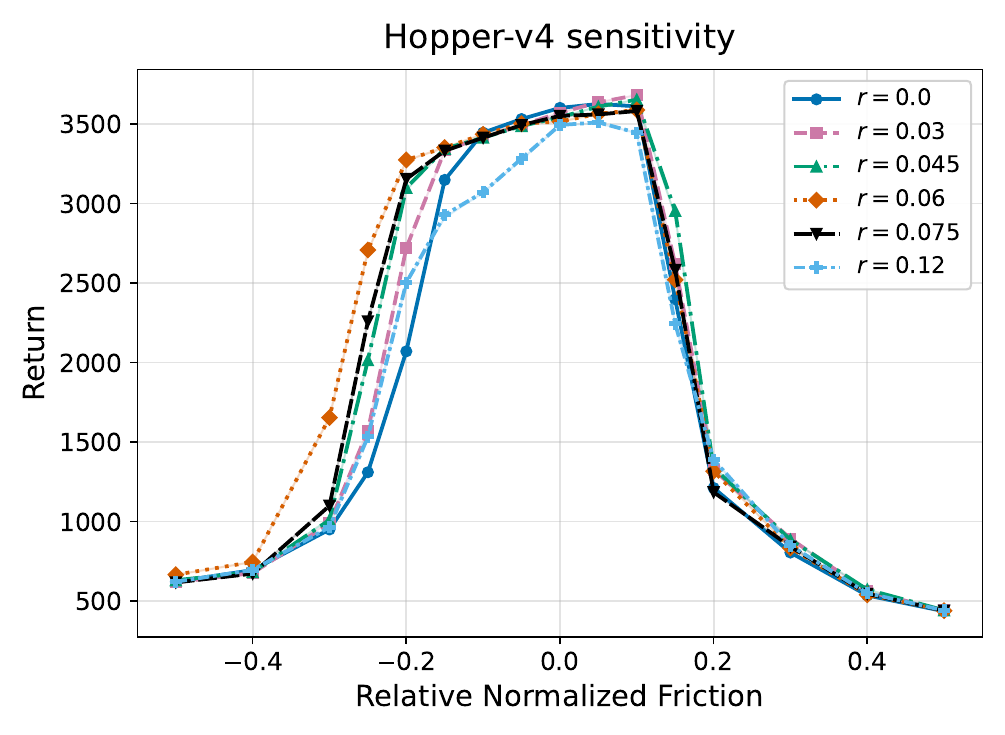}
        \caption{Uncertainty set radius.}
        \label{subfig:radius_ablation}
    \end{subfigure}
    \caption{Influence of the rollout depth and the uncertainty set radius $r^2$. Confidence intervals are omitted for visual clarity.}
\label{fig:ablations}
\end{figure}


\section{Implementation details and reproducibility.}

We tune the hyperparameters manually by considering that increasing $r$ causes more \gls{ood} usage of the transition model, while the treshold $h$ mitigates this. Therefore, we iteratively increase $r$ (starting at 0.01) until the nominal performance collapses, at which point we decrease the threshold $h$ to fix this. We keep iterating this process until a balance is found where no robustness is gained anymore. The amount of MPC iterations was straightforward to tune, as we can just look at after how many iterations the policy $\pi^{mpc}$ does not change meaningfully anymore, and use this amount of iterations as a cutoff. We often saw convergence after 5 iterations, hence we took 7 to introduce some margin. Finally, we observed that Reacher-v4 was insensitive to the rollout length $N$, and 5 was always enough in our observations. For Hopper-v4, increasing $N$ significantly increases the robustness, untill the rollouts become too long and compounding model error takes over. We observed an explosion of compounding model for $N>70$, hence we chose $N=40$ to introduce margin. All of the hyperparameters are listed in Table \ref{table:hparams}.

\begin{table}[h]
\centering
\begin{tabular}{l c c}
\hline
\textbf{Hyperparameter} & \textbf{Hopper-v4} & \textbf{Reacher-v4}\\
Rollout Depth $N$ & 40 & 5 \\ 
Threshold $h$ & 5.0 & 0.05 \\  
L2 disturbance size $r$ & 0.06 & 0.3 \\   
MPC iterations per step & 7 & 7 \\
\end{tabular}
\caption{Hyperparameters}
\label{table:hparams}
\end{table}

To implement MPC, we used the code provided by \cite{hansen2024tdmpc} and ported this to Jax for performance reasons. No modifications were made to this algorithm except the ones mentioned in Sec. \ref{sec:methodology}. The baseline \gls{mbpo} algorithm is exactly as described by \cite{janner2019trust}, including all hyperparameters. We used the same hyperparameters on Reacher-v4 as Hopper-v4, except that we found $15k$ environment steps to be enough, compared to $125k$ for Hopper-v4.

We base the evaluation environments on those provided by \cite{zhou2024natural}.



\end{document}